\documentclass{article}

\usepackage[preprint]{neurips_data_2024}

\usepackage[utf8]{inputenc} %
\usepackage[T1]{fontenc}    %
\usepackage{hyperref}       %
\usepackage{url}            %
\usepackage{booktabs}       %
\usepackage{nicefrac}       %
\usepackage{microtype}      %
\usepackage{xcolor}         %
\usepackage{epsfig}
\usepackage{graphicx}

\usepackage{amsmath,amsfonts,amsthm,amssymb}

\usepackage{pifont}
\usepackage{color}
\usepackage{wrapfig}
\usepackage{makecell}
\usepackage{colortbl}
\usepackage{multirow}
\usepackage{fixltx2e}
\usepackage{subcaption}
\usepackage{etoolbox}
\usepackage{multicol}
\usepackage{enumitem}
\usepackage{comment}

\definecolor{LightCyan}{rgb}{0.88,1,1}
\definecolor{mygray}{gray}{0.9}
\definecolor{mygray2}{gray}{0.6}

\title{MM-Instruct: Generated Visual Instructions for Large Multimodal Model Alignment}

\author{%
  Jihao Liu$^{1,2 \ *}$\quad Xin Huang$^{6,2 \ *}$ \quad Jinliang Zheng$^{5,2 \ *}$ \quad Boxiao Liu$^{2}$ \quad Jia Wang$^{6}$ \\ 
  \textbf{Osamu Yoshie$^{6}$ \quad Yu Liu$^{2}$ \quad Hongsheng Li$^{1,3,4}$ } \\
  $^1$CUHK MMLab \quad
  $^2$SenseTime Research  \\
  $^3$Shanghai AI Laboratory \quad
  $^4$CPII under InnoHK \\
  $^5$Institute for AI Industry Research (AIR), Tsinghua University
  $^6$Waseda University
}

\begin{document}

\maketitle

\begin{abstract}
    This paper introduces MM-Instruct, a large-scale dataset of diverse and high-quality visual instruction data designed to enhance the instruction-following capabilities of large multimodal models (LMMs).
    While existing visual instruction datasets often focus on question-answering, they struggle to generalize to broader application scenarios such as creative writing, summarization, or image analysis.
    To address these limitations, we propose a novel approach to constructing MM-Instruct that leverages the strong instruction-following capabilities of existing LLMs to generate novel visual instruction data from large-scale but conventional image captioning datasets.
    MM-Instruct first leverages ChatGPT to automatically generate diverse instructions from a small set of seed instructions through augmenting and summarization. 
    It then matches these instructions with images and uses an open-sourced large language model (LLM) to generate coherent answers to the instruction-image pairs. 
    The LLM is grounded by the detailed text descriptions of images in the whole answer generation process to guarantee the alignment of the instruction data.
    Moreover, we introduce a benchmark based on the generated instruction data to evaluate the instruction-following capabilities of existing LMMs. 
    We demonstrate the effectiveness of MM-Instruct by training a LLaVA-1.5 model on the generated data, denoted as LLaVA-Instruct, which exhibits significant improvements in instruction-following capabilities compared to LLaVA-1.5 models. 
    The MM-Instruct dataset, benchmark, and pre-trained models are available at \url{https://github.com/jihaonew/MM-Instruct}.
\end{abstract}

\section{Introduction}
\label{sec:intro}

Instruction-finetuned large language models (LLMs) have demonstrated superior capabilities for natural language tasks and real-world use cases. Inspired by the success of LLMs, finetuning large multimodal models (LMMs) with visual instruction data has attracted significant attention and made substantial progress in recent works~\cite{llava,llava_1_5,LVIS_Instruct4V,sharegpt4v}. Diverse and high-quality data plays an important role in the process of instruction finetuning. LLaVA~\cite{llava} first proposed generating visual instruction data with the assistance of GPT-4. Subsequent works, such as LVIS-Instruct4V~\cite{LVIS_Instruct4V} and ShareGPT4V~\cite{sharegpt4v}, further leveraged GPT-4V to produce visual instruction datasets.

\begin{figure}[t]
    \centering
    \includegraphics[width=1.0\linewidth]{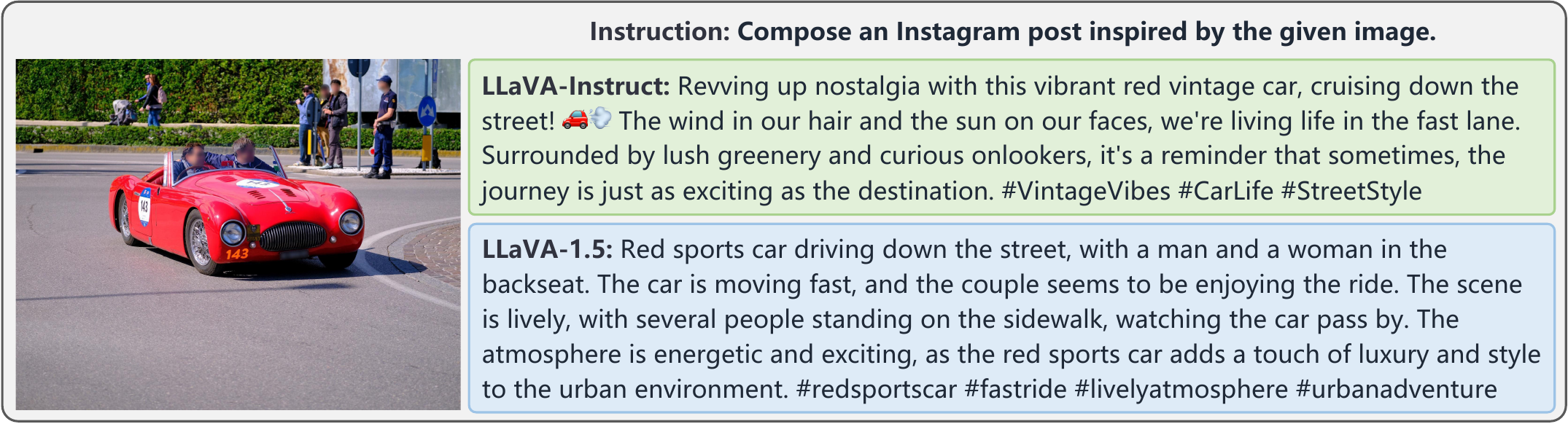}
    \vspace{-1.6em}
    \caption{\textbf{Example of instruction-following capability.} 
    For the given instruction, our baseline model (in green) follows the instruction and generates a post with engaging emojis and hashtags. In contrast, LLaVA-1.5's response describes a narrative instead of composing a post and has factual errors. This demonstrates our method is better able to comprehend and fulfill the intent of instructions.
    } 
    \vspace{-1.8em}
    \label{fig:instro_compare}
\end{figure}

While the visual instruction data generated by previous works have enabled the LMMs to better interact with humans on existing vision-language tasks, it poses significant challenges for applying LMMs to broader real-world use cases. In particular, most efforts~\cite{llava,sharegpt4v,LVIS_Instruct4V} focused on generating {\it question-answer} or {\it image-caption} pairs, which can significantly improve performance on existing benchmarks~\cite{goyal2017vqav2,marino2019okvqa,singh2019textvqa,fu2023mme}. However, these models often fail to follow users' requests in real-world scenarios like creative writing, summarization, or image analysis tasks that differ from established benchmarks, as demonstrated in Figure~\ref{fig:instro_compare}.
Manually collecting diverse data directly from real-world users mitigates these issues but requires substantial financial and human resources that are prohibitive for regular research groups to scale up.

To address these challenges, we introduce MM-Instruct, a novel dataset and benchmark specifically designed to enhance and evaluate the instruction-following capabilities of LMMs in real-world use cases. 
Directly generating high-quality and diverse instruction-tuning data at scale for training LMMs is difficult. However, we observe that a wealth of large-scale image captioning datasets already exist, providing numerous image-text pairs~\cite{gadre2024datacomp,schuhmann2022laion}. Despite their size and scope, these datasets often have textual descriptions that lack diversity, primarily focusing on basic image content descriptions. 
We recognize that generating diverse instructions and performing text-to-text answering are tasks where existing LLMs excel. 
This insight motivates our approach: leveraging the strong instruction-following capabilities of LLMs to construct MM-Instruct.
Our method consists of an automated pipeline with steps for instruction construction, instruction-following answer generation, and data filtering. Starting from limited seed instructions, we first prompt ChatGPT to generate diverse instructions guided by detailed image descriptions and in-context examples, which are then merged and summarized based on their similarities.
These instructions are matched with relevant images using a pre-trained CLIP model~\cite{clip}.
We then leverage the instruction-following capabilities of a powerful LLM to generate coherent answers to these instruction-image pairs, incorporating detailed image descriptions during the generation process to ensure that the answers are aligned with the images' contents and the paired instructions. 
We construct a large-scale visual instruction dataset by inferring an open-sourced LLM (e.g., Mixtral-8x7b~\cite{mistral}), and starting from 43 seed instructions, our method finally generates 293 diverse instructions and builds a dataset consisting of 234k high-quality visual instruction-answer pairs.
Leveraging the generated instruction data, we develop LLaVA-Instruct, an LMM based on the LLaVA-1.5 framework~\cite{llava_1_5}, to serve as a baseline model to evaluate the effectiveness of our dataset.

To evaluate the instruction-following capabilities of exiting LMMs, we build an automated evaluation pipeline with a held-out subset of our dataset and use GPT-4V~\cite{gpt4} as the judge to compare answers from LLaVA-Instruct to other state-of-the-art LMMs on each example and compute the overall win rates.
Empirically, we find that existing LMMs struggle to follow the given instructions and accomplish user requests, even though they perform well on traditional VQA benchmarks. In comparison, our LLaVA-Instruct demonstrates significantly improved instruction-following capabilities. 
According to the preference judgments of GPT-4V, LLaVA-Instruct-7B produces equally or more preferable responses in 72\% of the cases compared to LLaVA-1.5-7B.
Surprisingly, LLaVA-Instruct also enhances performance on traditional VQA benchmarks and outperforms LLaVA-1.5 on 9 of 12 tasks we evaluated, indicating that our dataset can also improve the general capabilities of LMMs.

\section{Methods}
\label{sec:method}

Generating visual instruction data is crucial for aligning large multimodal models (LMMs) with user intentions, enabling richer and more natural human-agent interactions. While a wealth of large-scale image captioning datasets exists~\cite{gadre2024datacomp,schuhmann2022laion}, the textual descriptions in these datasets often lack diversity, primarily focusing on basic image content descriptions. This limits their effectiveness in training LMMs for a broader range of real-world instruction-following scenarios, such as creative writing or summarization, where LMMs still struggle. To address this, we introduce MM-Instruct, a diverse and high-quality instruction dataset specifically designed to enhance the instruction-following capabilities of LMMs. 
Our approach leverages the remarkable instruction-following capabilities of large language models (LLMs) to generate diverse instructions and perform text-to-text answering, transforming conventional image captioning data into a rich source of visual instruction data.
In this section, we first present how to generate diverse instructions and then introduce the procedure of generating instances to these instructions. After that, we present the data filtering strategies for processing the generated instruction data. Finally, we introduce the baseline model LLaVA-Instruct trained with newly curated visual instruction data.

\begin{figure}[t]
    \centering
    \includegraphics[width=1.0\linewidth]{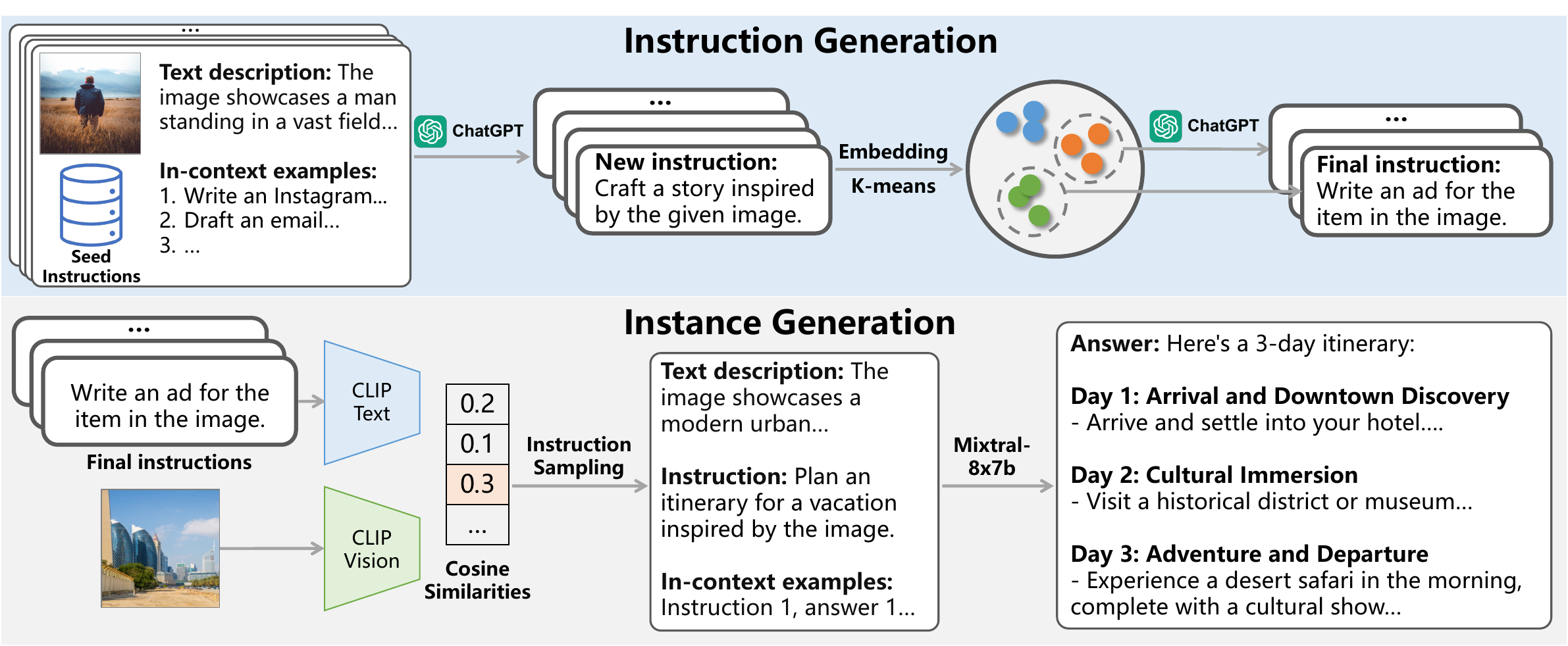}
    \vspace{-1.6em}
    \caption{\textbf{MM-Instruct for automatic instruction data generation.} (Top) In the \textbf{instruction generation} phase, ChatGPT is tasked with coming up with new instructions based on the image's text description. The generated instructions are then clustered and summarized into final instructions. (Bottom) In the \textbf{instance generation} phase, we first utilize CLIP to select a proper instruction for the input image and then employ Mixtral-8x7b to generate the answer adhering to the selected instruction.} 
    \vspace{-1em}
    \label{fig:pipeline}
\end{figure}

\begin{figure}[t]
    \centering
    \includegraphics[width=1.0\linewidth]{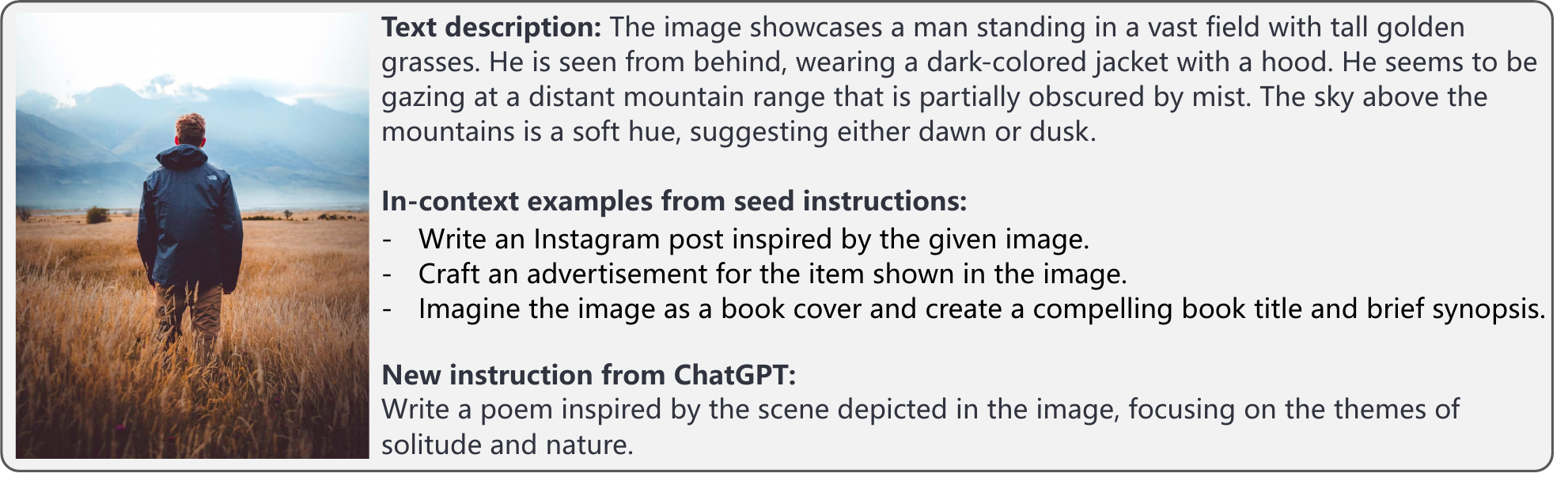}
    \vspace{-1.5em}
    \caption{\textbf{Illustration of instruction generation with in-context examples.} The text description is generated by an off-the-shelf LMM. The in-context examples are randomly sampled from 43 manually crafted seed instructions. We prompt ChatGPT to come up with a new instruction based on the text description and in-context examples.}
    \vspace{-1.5em} 
    \label{fig:instruction_generation}
\end{figure}

\subsection{Instruction Generation} 

Generating high-quality and diverse visual instructions poses unique challenges. 
In particular, it is challenging to develop creative ways to engage with image content through formulated tasks and novel perspectives. This requires imagination to conceptualize new scenarios and tasks linked to the image. Moreover, it requires accurately grounding generated instructions to the visual details within images, such as precisely describing depicted objects, actions, and relationships.
Facing these challenges, we leverage the powerful generative capabilities of LLMs (e.g., ChatGPT) to generate new instructions, while providing visual grounding through detailed image descriptions. Specifically, we first use an existing LMM~\cite{wang2023cogvlm} to generate detailed textual descriptions of objects, actions, and contexts depicted in input images. We then utilize ChatGPT to craft new instructions for each image based on its description. To guide the generation process, we also provide ChatGPT with in-context instruction examples from a manually curated seed collection. By conditioning generation on both the image description and related instruction samples, our approach balances creativity with adherence to visual details and consistency across examples. The overall process of instance generation is illustrated in Figure~\ref{fig:pipeline} (top).
We also present an example in Figure~\ref{fig:instruction_generation} to detail the process.

We generate about 50k initial instructions, which cover different topics or commands. However, we also identify two key issues among the initial ones. First, we observe duplications among the generated instructions, with ChatGPT producing similar or identical responses for different input images. Second, many instructions are overly specific, referencing details like product names that limit their reusability for other related images. To address these issues, we leverage clustering to consolidate the instructions. 
For the sake of simplicity, we utilize the $k$-means algorithm~\cite{k-means} with $k$ being heuristically set to 300 for performing clustering on the embeddings of the initial 50k instructions to group semantically similar instructions together. 
We then prompt ChatGPT to merge each cluster into a single, consolidated instruction. 
This refinement process helps to produce a more generalized and less duplicated set of instructions while maintaining the core tasks represented in the original instructions. 
Through this process, we obtained a final set of around 300 instructions with diversity and high reusability for new image data.

\begin{figure}[t]
    \centering
    \includegraphics[width=0.8\linewidth]{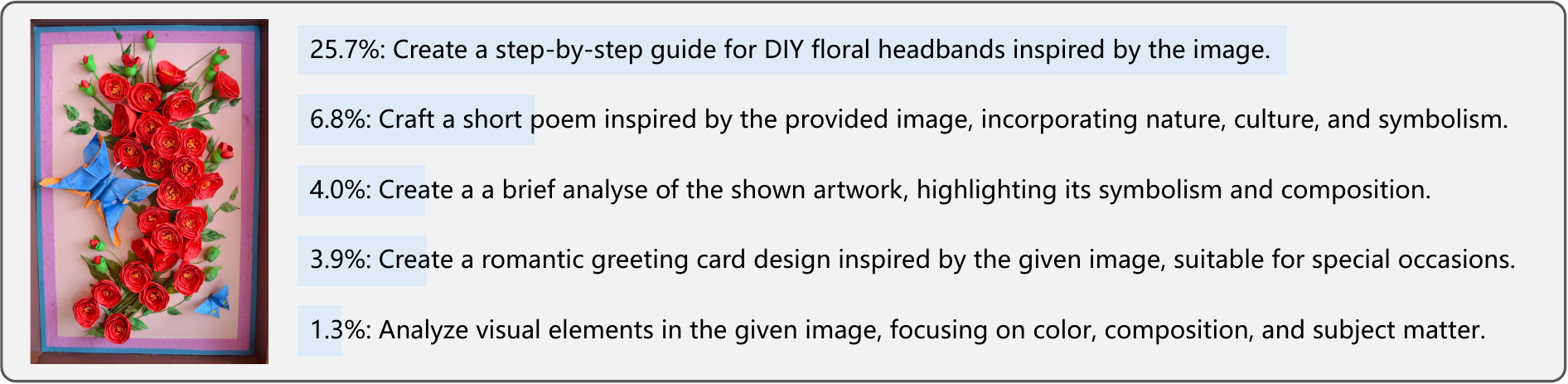}
    \vspace{-0.5em}
    \caption{\textbf{Example of image-instruction matching.} We show the top 5 instructions that match the example image, along with their corresponding scores. } 
    \vspace{-1.5em} 
    \label{fig:image_instruct_match}
\end{figure}

\subsection{Instance Generation}
Given the new instructions, we subsequently generate instances (i.e., image-instruction-answer triplets) for our final training. Generating a large-scale instance dataset involves two major problems: (1) how to pair suitable instructions with random images, and (2) how to generate high-quality answers to the selected instructions. To tackle these problems, we employ a pretrained CLIP model~\cite{clip} for image-instruction matching and leverage an off-the-shelf LLM~\cite{mistral} to generate accurate and coherent answers.
We illustrate the overall process in Figure~\ref{fig:pipeline} (bottom).

To begin, we calculate the cosine similarities between the CLIP image embedding of a given image and the CLIP text embeddings of all available instructions, with an example illustrated in Figure~\ref{fig:image_instruct_match}.
Using multinomial sampling with the cosine similarities as weights, we sample a specific instruction for the given image.
This instruction is then provided to an LLM alongside a detailed text description of the image, ensuring that the generated answer remains aligned with the visual context, to generate the final answer.
To strike a balance between the answer quality and total cost, we adopt a two-stage approach for answer generation. Initially, we leverage a more powerful language model, such as GPT-4, to generate multiple examples for each instruction in stage 1. These examples serve as in-context samples for the follow-up generation. Subsequently, we employ an open-sourced LLM, such as Mistral-8x7b~\cite{mistral}, for large-scale stage-2 generation, utilizing the GPT-4's generated in-context examples to enhance the quality of the final generated answers.

\subsection{Data Filtering}
To ensure the high quality of our instruction data at a large scale, we employ a series of heuristics to effectively filter out low-quality instances. 
First, we preprocess the source data by removing images with incomplete or overly short captions, as well as images with widths or heights under 100 pixels.
For the generated instances, we remove samples with inappropriate instructions by asking Mistral-8x7b whether an image's text description matches the selected instruction. Additionally, we develop heuristic rules to identify and filter instances that exhibit undesirable characteristics, such as discarding samples with incomplete answers or those that contain invalid repeat patterns.

\subsection{LLaVA-Instruct}
We instruction-tune an LMM (e.g., LLaVA-1.5~\cite{llava_1_5}) as a baseline model for MM-Instruct by training on the data, which is named {\it LLaVA-Instruct}. 
We transform our instruction data into the following format for training,
\[
\small \nonumber
    \texttt{X}_{\texttt{sys}}\texttt{<SEP>}
    \texttt{USER: } \texttt{<img>}\texttt{\textbackslash n}\texttt{X}_{\texttt{ins}}\texttt{<SEP>}
    \texttt{ASSISTANT: } \textcolor{blue}{\texttt{X}_{\texttt{ans}}\texttt{</s>}}
\]
where $\texttt{X}_{\texttt{sys}}$, $\texttt{X}_{\texttt{ins}}$, and $\texttt{X}_{\texttt{ans}}$ denote system message, instruction, and answer. The separation symbol $\texttt{<SEP>}$ is set as a single space aligning the LLM's setting. The $\texttt{<img>}$ placeholder is replaced with the image's embeddings.
Note that only the output of the assistant (text in blue) is used to compute the loss in the auto-regressive model. We follow the implementation of LLaVA-1.5 and do not perform specific optimization for LLaVA-Instruct. We leave the training details in supplementary materials.

\subsection{Benchmark}
\label{sec:eval_instruct}

Despite the comprehensive evaluation of existing vision-language benchmarks~\cite{goyal2017vqav2,marino2019okvqa,singh2019textvqa}, these benchmarks primarily focus on assessing the perception capabilities of LMMs. 
In order to thoroughly examine the instruction-following capabilities of LMMs, we create a new test set and employ the state-of-the-art LMM GPT-4V~\cite{gpt4} as the judge for performance evaluation. Specifically, we withhold 33 instructions and manually select 3 proper images for each instruction. We then employ GPT-4 to generate the target answer for each instruction-image pair. Each instance is manually checked to ensure the data quality. When comparing our baseline model to other models, we generate a single answer for each instance and prompt GPT-4V to compare LLaVA-Instruct's outputs to other models' and label which one it prefers. We calculate the overall win rate for performance comparison.

\section{Experiments}
\label{sec:exp}

We conduct extensive experiments and analysis to evaluate the effectiveness of our dataset.
We first describe our experimental setups and then present our baseline model's results on standard vision-language benchmarks. We also assess the instruction-following capabilities of LMMs and the quality of our generated instruction data. Finally, we perform ablation studies to examine the design choice of our method.

\subsection{Experimental Setups}

\noindent\textbf{Data.}\quad We generate instruction-tuning data from two datasets: Segment Anything 1 Billion (SA-1B)~\cite{kirillov2023segment} and DataComp-1B~\cite{gadre2024datacomp}. Specifically, we randomly sample 400k images from each dataset and use the off-the-shelf CogVLM model~\cite{wang2023cogvlm} to caption the images with the prompt ``Describe the image in detail''. For images from DataComp-1B, we also utilize the CapsFusion model~\cite{yu2023capsfusion} to merge the original caption with the generated caption. Finally, 234k instruction-tuning data are obtained after the data filtering. We combine our generated data with the original data from LLaVA-1.5~\cite{llava_1_5} for instruction-tuning the baseline model.

\noindent\textbf{Baseline model.}\quad Following LLaVA-1.5~\cite{llava_1_5}, we employ CLIP ViT-L/336px~\cite{clip} to encode images and Vicuna v1.5 7B/13B~\cite{vicuna} for text encoding. 
The vision-language connector is directly inherited from LLaVA-1.5 since our generated data is primarily used for instruction-tuning.

\noindent\textbf{Hyperparameters.}\quad We employ 3 and 2 in-context examples for instruction generation and instance generation respectively. To perform clustering, we extract embeddings of the instructions with the off-the-shelf Sentence-BERT model~\cite{sentence-bert} and conduct $k$-means~\cite{k-means} clustering with $k$ being empirically set to 300. For instruction-image matching, we utilize the off-the-shelf CLIP ViT-L model~\cite{clip} to calculate the cosine similarities. To enable fair comparison, we follow the original LLaVA-1.5's hyperparameters to instruction-tune our LLaVA-Instruct model. LLaVA-Instruct is trained on the combined dataset for 1 epoch. All the used prompts are detailed in the supplementary materials.

\noindent\textbf{Evaluation.}\quad We employ two zero-shot settings for performance evaluation. First, we conduct comprehensive evaluations on 12 vision-language benchmarks as demonstrated in Table~\ref{tab:sota_comparison}. These benchmarks cover a broad range of tasks to examine different capabilities of LMMs, such as captioning, mathematics, and spatial reasoning.
Additionally, we evaluate the instruction-following capabilities of LMMs on our proposed benchmark, as presented in Section~\ref{sec:eval_instruct}.

\begin{table}[t]
\setlength{\tabcolsep}{3pt}
\centering
\caption{
Comparison with state-of-the-art LMMs on 12 vision-language benchmarks. Our models consistently outperform LLaVA-1.5 using the same prompts and the same base LLM, demonstrating that the improved instruction-following capabilities are also beneficial for visual perception. 
Res indicates input image resolution.
We mark the best performance \textbf{bold} and the second-best \underline{underlined}.  
Benchmark names are abbreviated due to space limits. VQA-v2~\cite{goyal2017vqav2}; GQA~\cite{hudson2019gqa}; VizWiz~\cite{gurari2018vizwiz}; SQA$^\text{I}$: ScienceQA-IMG~\cite{lu2022learn}; VQA$^\text{T}$: TextVQA~\cite{singh2019textvqa}; POPE~\cite{li2023pope}; MME~\cite{fu2023mme}; MMB: MMBench~\cite{liu2023mmbench}; MMB$^\text{CN}$: MMBench-Chinese~\cite{liu2023mmbench}; SEED$^\text{I}$: SEED-Bench-Image~\cite{li2023seed}; LLaVA$^\text{W}$: LLaVA-Bench (In-the-Wild)~\cite{llava}; MM-Vet~\cite{yu2023mmvet}. $^*$The training images of the datasets are observed during training. 
}
\resizebox{1.0\linewidth}{!}{
\begin{tabular}{l llll | lllll | llllll l }
\toprule
Method & LLM & Res. & VQA$^\text{v2}$ & GQA & VizWiz & SQA$^\text{I}$ & VQA$^\text{T}$ & POPE & MME & MMB & MMB$^\text{CN}$ & SEED$^\text{I}$ & LLaVA$^\text{W}$ & MM-Vet \\
\midrule
BLIP-2~\cite{li2023blip} & Vicuna-13B & - & 41.0 & 41 & 19.6 & 61 & 42.5 & 85.3 & 1293.8 & -- & -- & -- & 38.1 & 22.4 \\
InstructBLIP~\cite{Dai2023InstructBLIP} & Vicuna-7B & 224 & -- & 49.2 & 34.5 & 60.5 & 50.1 & -- & -- & 36 & 23.7 & 58.8 & 60.9 & 26.2 \\
InstructBLIP~\cite{Dai2023InstructBLIP} & Vicuna-13B & 224 & -- & 49.5 & 33.4 & 63.1 & 50.7 & 78.9 & 1212.8 & -- & -- & -- & 58.2 & 25.6 \\
Shikra~\cite{chen2023shikra} & Vicuna-13B & 224 & 77.4$^*$ & -- & -- & -- & -- & -- & -- & 58.8 & -- & -- & -- & -- \\
IDEFICS-9B~\cite{idefics} & LLaMA-7B & 224 & 50.9 & 38.4 & 35.5 & -- & 25.9 & -- & -- & 48.2 & 25.2 & 44.5 & -- & -- \\
IDEFICS-80B~\cite{idefics} & LLaMA-65B & 224 & 60.0 & 45.2 & 36.0 & -- & 30.9 & -- & -- & 54.5 & 38.1 & -- & -- & -- \\
Qwen-VL~\cite{bai2023qwen} & Qwen-7B & 448 & 78.8$^*$ & 59.3$^*$ & 35.2 & 67.1 & 63.8 & -- & -- & 38.2 & 7.4 & 62.3 & -- & -- \\
Qwen-VL-Chat~\cite{bai2023qwen} & Qwen-7B & 448 & 78.2$^*$ & 57.5$^*$ & 38.9 & 68.2 & \underline{61.5} & -- & 1487.5 & 60.6 & 56.7 & 65.4 & -- & -- \\
LLaVA-1.5~\cite{llava_1_5} & Vicuna-1.5-7B & 336 & 78.5$^*$ & 62.0$^*$ & 50.0 & 66.8 & 58.2 & 85.9 & 1510.7 & 64.3 & 58.3 & -- & 63.4 & 30.5 \\ 
LLaVA-1.5~\cite{llava_1_5} & Vicuna-1.5-13B & 336 & \underline{80.0}$^*$ & \underline{63.3}$^*$ & \underline{53.6} & \textbf{71.6} & 61.3 & 85.9 & \underline{1531.3} & \underline{67.7} & \textbf{63.6} & \underline{68.2} & \textbf{70.7} & \underline{35.4} \\ \midrule
\rowcolor{black!10}
LLaVA-Instruct & Vicuna-1.5-7B & 336 & 79.3$^*$ & 62.7$^*$ & 50.2 & 70.0 & 58.7 & \underline{86.8} & 1523.8 & 66.4 & 60.0 & 67.7 & 65.1 & 32.9 \\
\rowcolor{black!10}
LLaVA-Instruct & Vicuna-1.5-13B & 336 & \textbf{80.3}$^*$ & \textbf{63.8}$^*$ & \textbf{56.3} & \underline{70.8} & \textbf{61.8} & \textbf{87.0} & \textbf{1569.7} & \textbf{67.9} & \underline{62.5} & \textbf{69.4} & \underline{70.1} & \textbf{37.1} \\
\bottomrule
\end{tabular}
}
\vspace{-1em} 
\label{tab:sota_comparison}
\end{table}

\begin{figure}[t]
    \centering
    \includegraphics[width=0.8\linewidth]{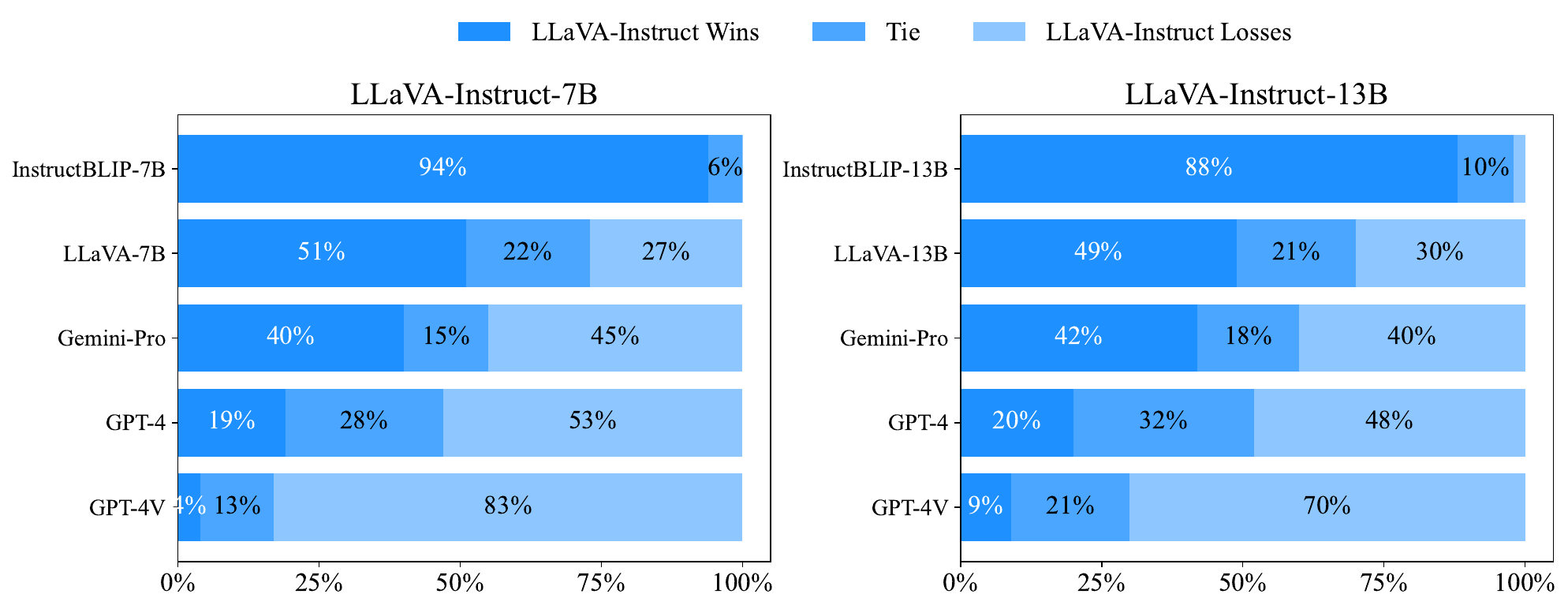}
    \vspace{-0.8em} 
    \caption{\textbf{Instruction-following evaluation using GPT-4V as the judge.} We compare LLaVA-Instruct-7B/13B to 5 different approaches. Our baseline models demonstrate stronger instruction-following capabilities than InstructBLIP or LLaVA under the same model sizes.}
    \vspace{-0.8em} 
    \label{fig:win_rate}
\end{figure}

\begin{figure}[t]
    \centering
    \includegraphics[width=0.7\linewidth]{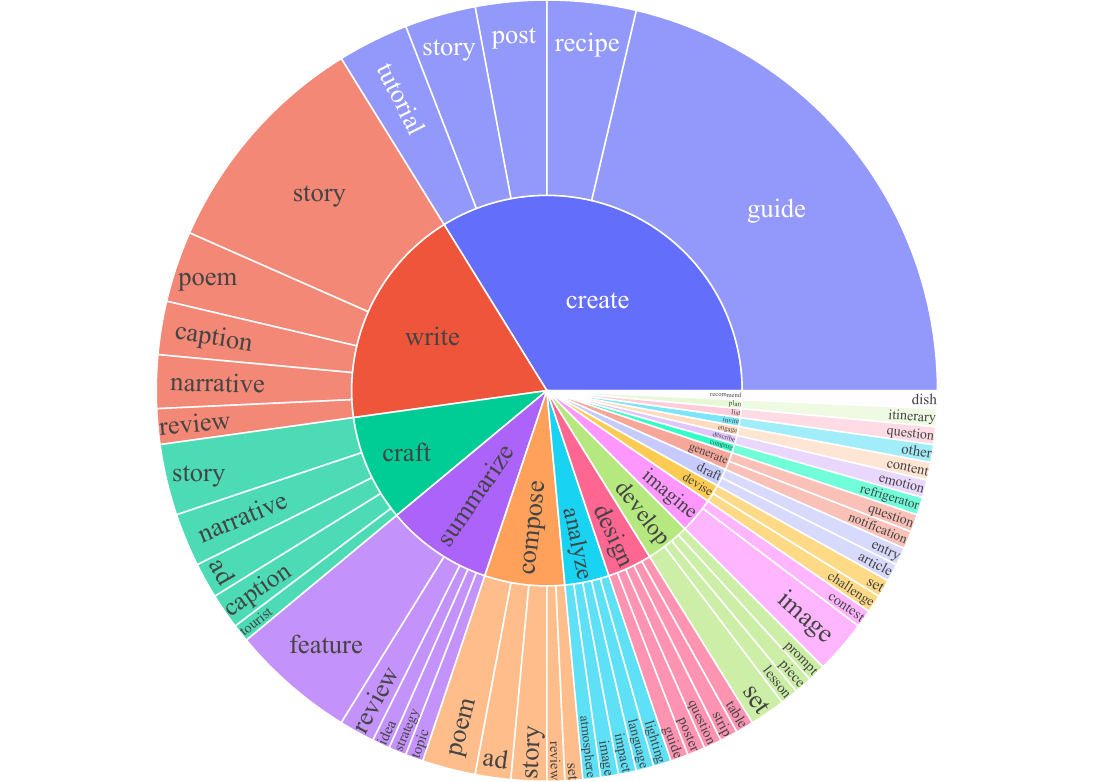}
    \caption{\textbf{Illustrate of instruction diversity.} We show the top 20 most common root verbs (inner circle) and their top 5 direct noun objects (outer circle) in our generated instructions. These instructions cover a broad of topics in real-world scenarios.} 
    \vspace{-1.5em} 
    \label{fig:data_diversity}
\end{figure}

\begin{figure}[t]
    \centering
    \includegraphics[width=1.0\linewidth]{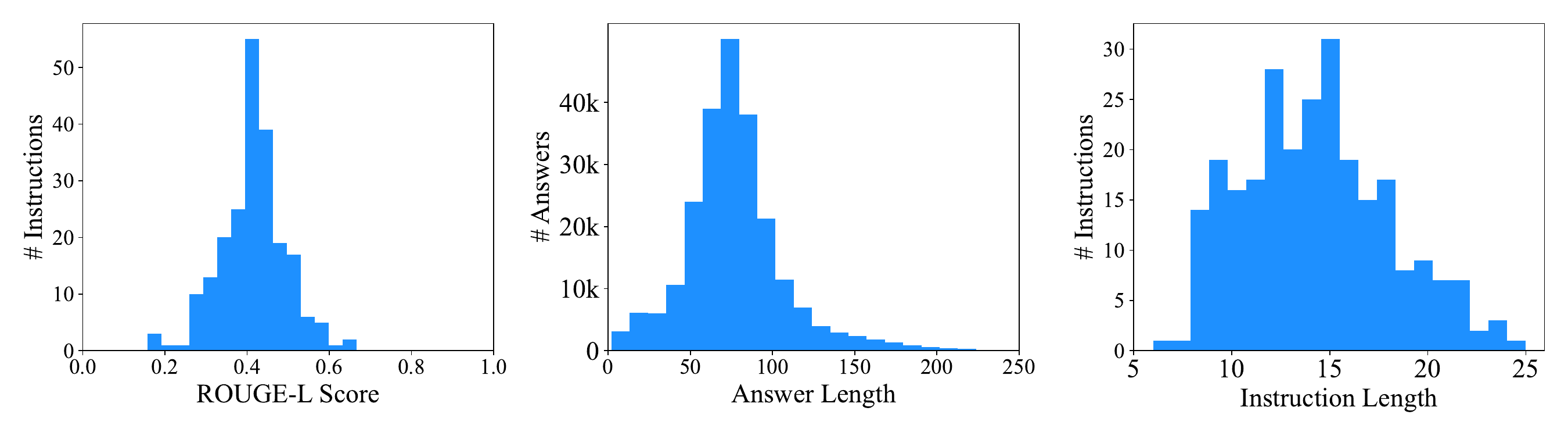}
    \vspace{-2em} 
    \caption{\textbf{Statistics of instructions and answers.} (Left) Distribution of ROUGE-L scores between generated instructions and their most similar seed instructions. (Middle) Distribution of answer lengths. (Right) Distribution of instruction lengths.} 
    \vspace{-1.5em} 
    \label{fig:statistics_data}
\end{figure}

\subsection{Performance on Vision-Language Benchmarks}
We conduct comprehensive comparisons with state-of-the-art LMMs on 12 vision-language benchmarks, with results shown in Table~\ref{tab:sota_comparison}. Compared to existing models like LLaVA-1.5~\cite{llava_1_5}, our baseline models achieve consistent improvements over most datasets across different model sizes using the same evaluation prompts and base LLM. 
Notably, our LLaVA-Instruct-13B outperforms LLaVA-1.5-13B on VizWiz~\cite{gurari2018vizwiz} and MME~\cite{fu2023mme} benchmarks by a large margin. 
Importantly, our approach does not rely on generating question-answer pairs to fit these benchmarks. Instead, all improvements are achieved solely through our diverse and high-quality instruction data, demonstrating that enhanced instruction-following capabilities can also lead to stronger visual perception abilities.

\subsection{Evaluation of Instruction-Following Capability}

We examine the instruction-following capabilities of existing LMMs and our LLaVA-Instruct on the benchmark introduced in Section~\ref{sec:eval_instruct}.
As shown in Figure~\ref{fig:win_rate}, our LLaVA-Instruct outperforms LLaVA-1.5~\cite{llava_1_5} or InstructBLIP~\cite{Dai2023InstructBLIP} significantly under the same model sizes, demonstrating that our generated data can effectively improve the instruction-following capabilities of LMMs. 
Moreover, against the strong Gemini-Pro model~\cite{team2023gemini}, LLaVA-Instruct-13B produces equally or more preferable responses in 60\% of the cases despite using a much weaker training protocol. 
We also observe that the win rate notably increases with a larger base LLM, highlighting the importance of the base LLM scale for instruction following.

\begin{figure}[t]
    \centering
    \includegraphics[width=0.75\linewidth]{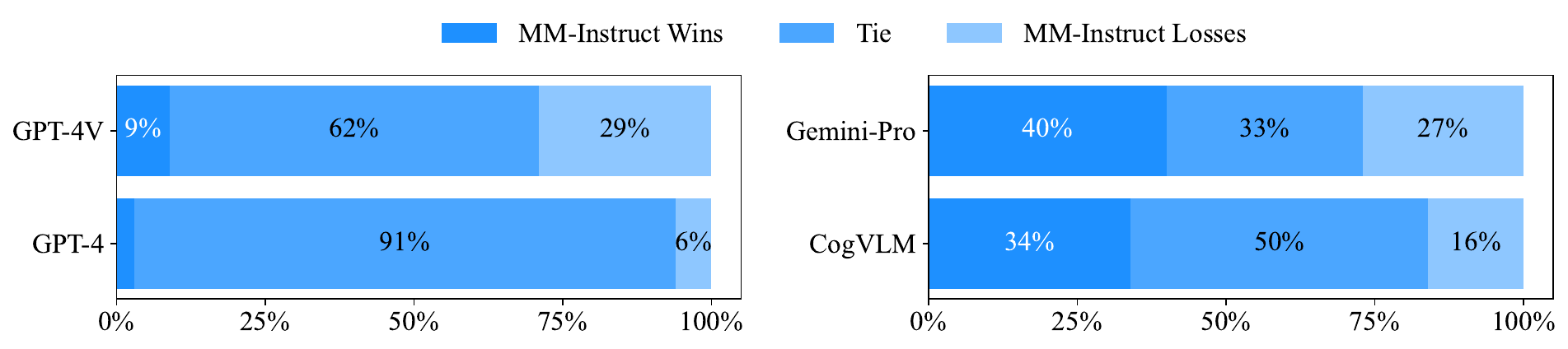}
    \vspace{-0.9em} 
    \caption{\textbf{Data quality examination using GPT-4V as the judge.} We randomly sample 100 instances from our generated data and compare their answers to those produced by various approaches. Our method can generate high-quality answers without relying on distillation from GPT-4V.} 
    \vspace{-1em} 
    \label{fig:data_quality}
\end{figure}

\begin{figure}[t]
    \centering
    \includegraphics[width=0.75\linewidth]{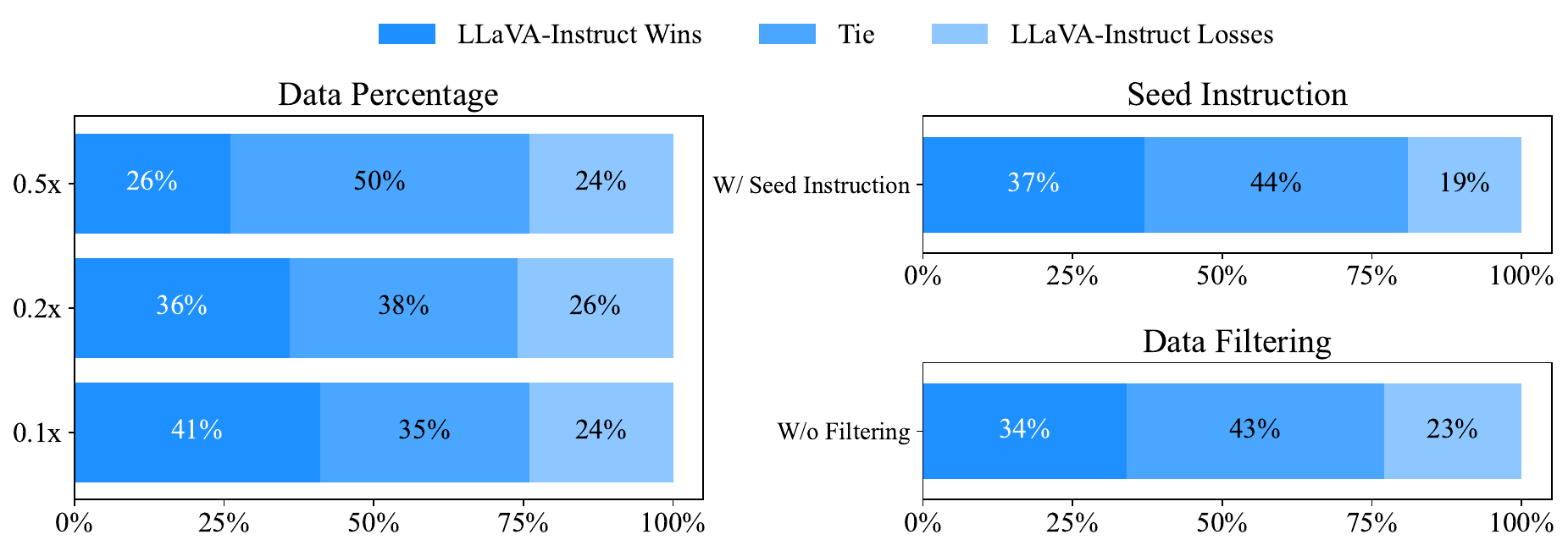}
    \vspace{-0.9em} 
    \caption{\textbf{Ablation study results.} We compare the instruction-following capabilities of our LLaVA-Instruct-7B model and models under different ablation settings. (Left) Impact of the data sizes used for instruction-tuning. (Top right) Comparison of using generated new instructions versus only seed instructions. (Bottom right) Impact of utilizing data filtering. }
    \vspace{-1em} 
    \label{fig:ablation}
\end{figure}

\subsection{Data Diversity and Data Quality}

\noindent\textbf{Data Diversity.} Following prior work~\cite{wang2022self}, we examine the data diversity by analyzing verb-noun structure in our generated instructions. Specifically, we parse the instructions using the Berkeley Neural Parser~\cite{kitaev2018constituency,kitaev2018multilingual} to extract the root verb and its direct noun object for each instruction. The top 20 most common root verbs and their top 5 noun objects are illustrated in Figure~\ref{fig:data_diversity}. 
We observe that our generated instructions cover a broad range of topics and formats in real-world scenarios, demonstrating exceptional diversity.
Moreover, we examine the difference between our generated instructions and our manually crafted seed instructions using the ROUGE-L~\cite{lin2004rouge} scores.
As shown in Figure~\ref{fig:statistics_data} (left), we observe low ROUGE-L scores between each generated instruction and its most similar seed instruction, indicating that our approach generates new instructions beyond the seed instructions. Figure~\ref{fig:statistics_data} also shows the distribution of answer and instruction lengths, illustrating the diversities in our generated instructions and answers.

\noindent\textbf{Data Quality.} To investigate the quality of the generated instruction data, we randomly sample 100 instances and compare our generated answers with state-of-the-art LMMs' answers, such as Gemini-Pro~\cite{team2023gemini} or GPT-4V~\cite{gpt4}. We employ GPT-4V to judge the answers and compute the overall win rate.
As shown in Figure~\ref{fig:data_quality}, our approach can generate high-quality answers against other strong LMMs. 
For example, we can produce equally or more preferable responses in 73\% and 71\% of the cases compared to Gemini-Pro and GPT-4V respectively.
Unlike existing approaches~\cite{sharegpt4v,LVIS_Instruct4V} that rely on distilling from GPT-4V, our method leverages open-source LLMs (e.g., Mixtral-8x7b) to generate target answers. These LLMs can produce high-quality answers while being more accessible and cost-effective.

\subsection{Ablation Studies}

We perform ablation studies to analyze the impact of different design choices on the instruction-following capabilities using the same evaluation method as in Section~\ref{sec:eval_instruct}. We show the comparison between LLaVA-Instruct-7B and models under different ablation settings in Figure~\ref{fig:ablation}. 
First, we study the effect of data sizes used for finetuning. As shown in Figure~\ref{fig:ablation} (left), we observe that increasing the instruction data size from 10\% to 20\% to 50\% of the original data size can lead to progressively better performance, showing that more data enables better learning of instruction-following.

To examine the benefits of using newly generated instructions, we build an additional dataset using only seed instructions. As shown in Figure~\ref{fig:ablation} (top right), using only seed instructions can lead to a significant performance drop, which shows the effectiveness of our generated new instructions. We also study the impact of data filtering on performance in Figure~\ref{fig:ablation} (bottom right). We observe that conducting data filtering can effectively improve the instruction-following capabilities, highlighting the importance of high-quality instruction data.

\subsection{Qualitative Results}

\begin{figure}[t]
    \centering
    \includegraphics[width=0.9\linewidth]{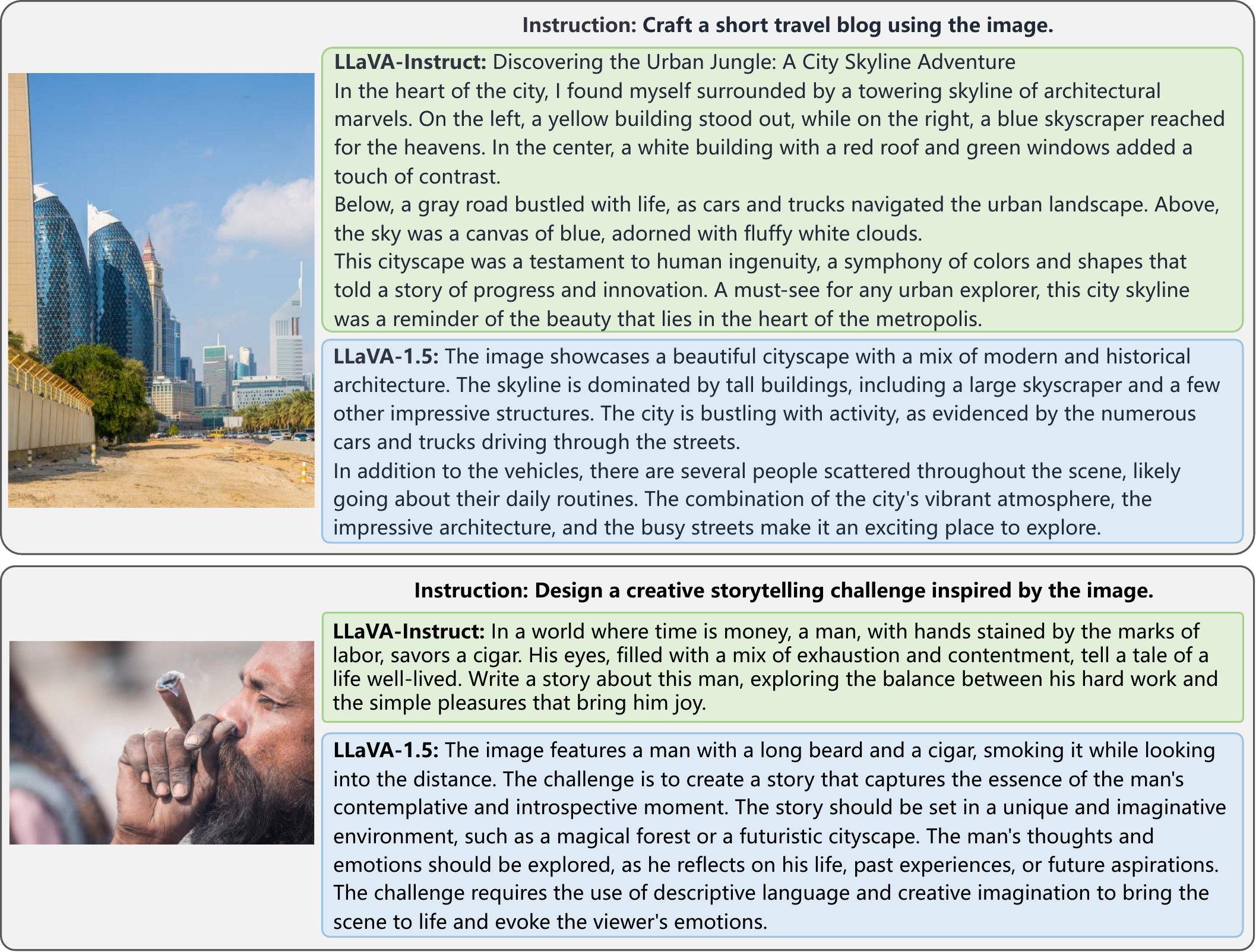}
    \vspace{-0.6em} 
    \caption{\textbf{Illustration of better instruction following.} LLaVA-Instruct can better capture the user's intent and give more coherent answers.}
    \vspace{-1em} 
    \label{fig:mix_exp}
\end{figure}

This section qualitatively analyzes how LLaVA-Instruct improves upon LLaVA-1.5 in comprehending and fulfilling visual instructions. Note that all the studied cases are not observed by LLaVA-Instruct during the finetuning stage.
Figure~\ref{fig:mix_exp} (top) illustrates examples where LLaVA-Instruct is better able to follow the given instruction compared to LLaVA-1.5, which struggles and provides a factual description rather than a response aligned with the instruction's intent.
Moreover, example in Figure~\ref{fig:mix_exp} (bottom) demonstrates that while both models may understand the instruction, LLaVA-Instruct can provide responses that are more coherent and creative. For instance, when instructed to ``Design a creative storytelling challenge inspired by the image.'', LLaVA-Instruct starts with ``In a world where time is money'' and creates a unique semantics. 
These qualitative comparisons suggest that LLaVA-Instruct has gained a stronger ability to align its outputs with the tasks implied by different types of instructions, demonstrating the effectiveness of finetuning on our generated instruction data.

\section{Related Works}
\label{sec:related}

Our work intersects with several active research areas, including the development of large multimodal models (LMMs)~\cite{llava,llava_1_5,li2023blip,Dai2023InstructBLIP,bai2023qwen,alayrac2022flamingo} instruction finetuning for both language and vision-language tasks~\cite{mishra2021instruction-follow,ouyang2022instruction-follow,flan,flan-t5}, and the generation of instruction data, particularly in the visual domain~\cite{sharegpt4v,LVIS_Instruct4V,li2023otter,li2023m,xu2022multiinstruct,gong2023multimodal,li2023textbind}. While prior work has made significant strides in each of these areas, existing visual instruction datasets often focus on question-answering formats~\cite{sharegpt4v,LVIS_Instruct4V} and struggle to generalize to the diverse range of real-world use cases we aim to address with MM-Instruct. Due to space limitations, we provide a more detailed discussion of related work in the supplementary materials.

\section{Conclusion}
\label{sec:conclusion}

In this paper,  we present MM-Instruct, a novel dataset and benchmark designed to enhance and evaluate the instruction-following capabilities of LMMs for real-world applications. By leveraging the strengths of existing LLMs, we transform conventional image-captioning data into a diverse and rich source of visual instruction data. Our experiments demonstrate that our baseline model LLaVA-Instruct, trained on MM-Instruct, exhibits significantly improved instruction-following abilities compared to existing LMMs, particularly in scenarios beyond traditional benchmarks.


\end{document}